# A Survey on Artificial Intelligence and Data Mining for MOOCs

Simon Fauvel, Han Yu

Nanyang Technological University, Singapore

**Abstract**

Massive Open Online Courses (MOOCs) have gained tremendous popularity in the last few years. Thanks to MOOCs, millions of learners from all over the world have taken thousands of high-quality courses for free. Putting together an excellent MOOC ecosystem is a multidisciplinary endeavour that requires contributions from many different fields. Artificial intelligence (AI) and data mining (DM) are two such fields that have played a significant role in making MOOCs what they are today. By exploiting the vast amount of data generated by learners engaging in MOOCs, DM improves our understanding of the MOOC ecosystem and enables MOOC practitioners to deliver better courses. Similarly, AI, supported by DM, can greatly improve student experience and learning outcomes. In this survey paper, we first review the state-of-the-art artificial intelligence and data mining research applied to MOOCs, emphasising the use of AI and DM tools and techniques to improve student engagement, learning outcomes, and our understanding of the MOOC ecosystem. We then offer an overview of key trends and important research to carry out in the fields of AI and DM so that MOOCs can reach their full potential.

## 1. Introduction

Humans are lifelong learners. From our first breaths to our last, we are constantly learning: we learn about the world around us, we learn about each other, we gather new knowledge. To help us in this never-ending task, we have set up both formal and informal education systems. Massive Open Online Courses (MOOCs) are one of the latest trends in education, making it easier for learners to access free, high-quality learning opportunities. Given a computer and an Internet connection, students around the world have open access to high-quality courses from the best schools and organisations of the planet. Targeting learners who do not have access to formal higher education as well as lifelong students looking to gain new skills and knowledge, MOOCs connect learners from all over the world.

Fuelled by improvements in cloud computing and widespread access to Internet, MOOCs have gained tremendous popularity in the last few years. The term 'MOOC' first appeared in 2008 but it's only in 2011 that it started to attract attention, when Stanford University launched three MOOCs that had a total enrolment of over 100,000 each [1]. Since then, growth has been fast. From thousands of learners and a handful of partner universities and available courses in 2011, MOOCs now attract millions of learners from all over the world, and most of the world's top institutions have joined the movement, often offering many courses each. Coursera, the biggest MOOC provider, boasts over 14,000,000 registered students, 1068 available courses, and 121 partner universities[1].

Online education is not a new idea; it appeared around the same time as Internet, and can be seen as an extension of distance learning in the digital age [2]. What, then, differentiates MOOCs from other online education systems? The scale, for one: traditional online education systems are built to accommodate tens of learners and would not be able to sustain the load of thousands of simultaneous students. But more importantly, MOOCs create a learning community that is typically not present in traditional online education systems (or at least not at that scale and depth), in which learners can communicate and interact with one another.

[1] Retrieved from https://www.coursera.org/ on 27 July 2015

There do not exist specific quantitative criteria to determine whether a given course or platform falls under the MOOC paradigm or not. We offer the following generic definition for MOOCs:

- Massive: they can accommodate thousands to tens of thousands of students taking the same course simultaneously;
- Open: they are freely accessible – anyone can register with no financial barriers, and there are no prerequisites or preconditions for learners;
- Online: they are accessed through a computing device (e.g. a computer, a smartphone or a tablet) with an Internet connection;
- Course: they offer a well thought-out learning sequence, as opposed to unconnected learning objects or modules.

MOOCs have generally been divided into two broad categories: connectivist MOOCs (cMOOCs), based on the connectivist pedagogy approach, and xMOOCs, that closely resemble traditional university courses. cMOOCs emphasise the creation of connections between learners, and the learning contents are not predefined; instead, they evolve according to the kinds of discussions happening between learners, which are facilitated by skilled instructors. xMOOC on the other hand generally have a predefined learning sequence; materials are prepared ahead of time by an instructor, and students go through them in the given sequence [3]. Today, the vast majority of available MOOCs are of the xMOOC kind.

MOOCs are distributed through online platforms. Popular platforms include Coursera[2], edX[3] and Udacity[4], but there are dozens of other available platforms, each with its particular characteristics and features. In general, platforms include the following three components: course contents, community building tools, and platform tools. Course contents can be divided into informational assets and interactive assets. Informational assets include videos (by far the main content delivery strategy in MOOCs) and supporting learning materials (such as reading materials from textbooks or website, lecture slides, lecture notes, topic outline, etc.) Interactive assets include exercises, quizzes and exams for students to complete as part of their assessment. Community building tools include asynchronous tools (i.e. they do not rely on learners using them simultaneously) such as forums, as well as synchronous tools (which require learners to share some space simultaneously) such as chat rooms and real-time group discussions. They may also include group work tools, and peer support tools. Different platforms will use different tools. Platform tools include searching and recommendation features, as well as learner authentication. Most platforms also provide an interface for instructors to organise their course contents, and some basic statistics and data visualisation tools to support them in teaching. This common structure for MOOC platform is summarised in Figure 1.

Despite their tremendous success, there are a number of issues that prevent MOOCs from reaching their full potential. Currently the main issue with MOOCs is the low completion rate of most courses.

---

[2] https://www.coursera.org/
[3] https://www.edx.org/
[4] https://www.udacity.com/

Indeed, completion rate rarely exceeds 10% for a course, which is very low especially as compared to traditional courses [4]. Students dropping out of MOOCs report many different reasons for their behaviour, including a lack of motivation, feelings of isolation, lack of interactivity, and poor time management among other things. For MOOCs to be truly successful, the dropout rate needs to be lowered significantly.

MOOCs also represent a big resource pull from instructors. Creating and delivering a MOOC is often a time consuming endeavour that can also be quite costly financially. Upfront, instructors need to plan the contents and create them. Traditional instructors might not be fully equipped to deliver a MOOC-format course, which means that they need to spend more time crafting their content. It is also often difficult for them to reuse existing materials from other sources, for various reasons (e.g. copyright issues, difficulty to locate these materials, lack of compatibility between formats, etc.) Running the course itself is time consuming, especially when it comes to monitoring the forums and answering students' questions and concerns. Many instructors launch themselves into MOOCs with little awareness of what is to come, which can lead to bad surprises.

Another concern is about the openness of MOOCs. While most MOOC platforms indeed offer their courses for free (although some of them charge a premium for extra features such as certification), many platforms are not fully accessible to all learners due to linguistic, technological and disability barriers.

Finally, something needs to be said about the financial sustainability of MOOCs. MOOCs are currently primarily funded and supported by leading universities but at some point, MOOCs will need to be self-sustainable through a sound business model. At this point, as far as we know, no MOOC platform is financially viable.

Much research has already taken place about MOOCs. One research area that has gained significant popularity in the last few years is how artificial intelligence (AI) and data mining (DM) can contribute to better understanding the MOOC ecosystem, and how they can contribute to improving it. In a nutshell, AI studies how to design intelligent machines and systems that analyse their environment and take actions that maximise their chances of success. AI borrows from various fields such as computer science, mathematics, and robotics. An AI-augmented MOOC platform can enable a better understanding of how learning happens in MOOCs as well as a more engaging learning experience for learners, thus improving their learning outcomes. The goal of data mining is to extract useful knowledge, information and patterns from seemingly unstructured data. To do so, it borrows tools, techniques and algorithms mainly from AI and statistics. MOOCs offer a very rich environment for DM given the large amount of data generated by learners using the platform but making sense of such data is not a trivial task. In some ways, all AI research in MOOC can be considered under DM since it relies on data to operate – the goal isn't to come up with new theoretical developments in AI but rather to use AI tools to solve a problem that generally involves big data.

Research in AI and DM applied to MOOCs has contributed to our current understanding of the MOOC ecosystem and has a big part to play in the success they have enjoyed so far. More importantly, they have a crucial role to play in solving the current issues experienced in MOOCs. In this paper, we review the state-of-the-art AI and data mining research for MOOCs, emphasising the use of AI and DM tools and techniques to improve student engagement, learning outcomes, and our understanding of the MOOC ecosystem. This survey paper will be useful for anyone interested in contributing to AI and DM for MOOCs, allowing them to quickly get a sense of the current state-of-the-art and see the existing gaps in research. We also hope that this paper will be valuable to MOOC researchers and practitioners outside AI and DM, through highlighting the key general findings derived from this research area as well as promising future trends.

The rest of this paper is organised as follows. Section 2 presents a taxonomy of the fields of artificial intelligence and data mining. Sections 3 to 6 discusses the state-of-the-art AI and DM research at it pertains to MOOCs, covering student understanding, course contents, community building and platform, respectively. Section 7 presents key trends and future research while section 8 concludes this paper.

## 2. A Taxonomy of Artificial Intelligence and Data Mining

Before diving into the actual research on AI and DM for MOOCs, it is useful to present a taxonomy of these two fields so that we can better understand which aspects MOOC research focuses on.

AI is a broad area that spans many different sub-fields. We borrow the taxonomy from Russell and Norvig's "Artificial Intelligence: A Modern Approach" reference book [5]. They define five main areas:

- **Problem Solving**: 1) searching (search algorithms and solutions), 2) game theory and incentive mechanisms, and 3) constraint satisfaction problems.
- **Knowledge, reasoning and planning**: 1) knowledge engineering (propositional logic, first-order logic, inference), 2) planning algorithms and solutions, and 3) knowledge representation (ontologies, taxonomies, semantics, reasoning)
- **Uncertain knowledge and reasoning**: 1) knowledge and reasoning under uncertainty (traditional and Bayesian probability statistics, probabilistic reasoning, decision making)
- **Learning**: 1) machine learning (supervised, unsupervised and reinforcement statistical learning; knowledge-based learning; probabilistic learning)
- **Communicating, perceiving and acting**: 1) communication algorithms and solutions (natural language processing), 2) perception algorithms and solutions (image and object recognition), and 3) robotics (perception, planning and moving)

DM mainly employs techniques from AI (machine learning primarily) as well as statistical tools. DM methods can be broadly separated into two categories:

- Predictive: 1) interpolation and sequential prediction, and 2) supervised learning
- Descriptive: 1) exploratory analysis, and 2) clustering

In the next sections, we will review the state-of-the-art research by emphasising which AI and DM fields are exploited.

## 3. Better Understanding Learners

MOOCs are centred on learners, and as such a key goal of MOOCs is to better understand these learners. There already exists a large body of AI and DM research for this purpose, which we break down into three aspects: 1) modelling engagement and learning behaviours, 2) modelling, predicting and influencing learners' achievement, and 3) modelling learners knowledge.

### 3.1. Modelling Engagement and Learning Behaviours

If one wishes to improve MOOC outcomes for students, it is first necessary to better understand how learners engage with MOOC resources and tools. By their virtue of being open, MOOCs attract a large range of learners with widely different backgrounds and motivations for taking a MOOC. As such, many different learning patterns emerge. Much research has already taken place to better understand and categorise these different learning styles. We categorise the work according to the data used to come up with the different learner categories.

#### 3.1.1. Based on activities on the MOOC platform

MOOC platforms collect all activities that the learners engage in, down to the click level. This offers an abundant amount of data that can be used to extract learning styles. In [6], authors use two activities for this purpose: viewing a lecture and handing in an assignment for credit. They then use a histogram approach to cluster learners. They developed a taxonomy of learner behaviour that comprises five styles of engagement: viewers, solvers, all-rounders, collectors, bystanders.

#### 3.1.2. Based on performance

One way to categorise MOOC learning styles is to cluster learners according to their performance in a course (based on course assessments). In [7], authors use the performance of learners in the sequence of activities in the course (videos and exercises) to cluster learners into three broad categories: lurkers, participants that do not complete the course, and participants that complete the course. The clustering method used is not specified and seems to be based on simple statistics.

#### 3.1.3. Based on forum behaviour

As mentioned earlier, most courses offer a forum on which learners can exchange with one another. This offers a rich data source that can be mined for patterns. In [8], qualitative content analysis of forum posts is first conducted, based on three dimensions: learning, dialogue acts, and topic. The coding is done manually. It is then followed by Bayesian Non-negative Matrix Factorisation (BNMF) to extract communities of learners based on the codes. Different learner communities are identified for 2 distinct sub-forums. In one sub-forum, 4 communities are identified: committed crowd engagers, discussion initiators, strategists, and individualists. In the other, 5 communities are identified: instrumental help seekers, careful assessors, community builders, focused achievers, and project support seekers.

#### 3.1.4. Based on learners' stated intentions

Most MOOCs ask learners to fill in a pre-course survey that covers various elements such as background, expectations, and intentions. When such data is collected, it is possible to link learners' intentions with their learning behaviour in the MOOC. In [9], authors use this survey data to map it to learner categories. They identified four categories: no-shows, observers, casual learners, completers. The clustering is based on simple mappings between intentions and styles.

#### 3.1.5. Summary

As can be noted from the above discussions, different learning styles emerge based on analysing learners' interactions with MOOC resources and tools. One important element to note is that each research project comes up with its own categorisation of these learning styles, based on their specific purpose (since the end goal is generally not to simply identify learning styles but rather to infer other elements such as achievement). While there is value in this (and a point can be made that these different learning styles are generally quite similar and potentially equivalent), the discrepancy prevents comparisons between the works, which is not ideal.

### 3.2. Modelling, Predicting and influencing Learners' Achievement

After modelling engagement and learning behaviours, the next logical step is to better understand what leads to learner achievement (or, in other words, completing a MOOC). As mentioned in the introduction, one of the main issues with MOOCs currently is their low completion rate. As such, it should not come as a surprise that a significant amount of research has gone into looking at modelling, predicting and influencing learners' achievement, so that dropout rates can be reduced. The goal is usually to predict drop out as early as possible, so that remedial actions can be taken. We categorise the work according to the data used to model, predict, and influence achievement.

#### 3.2.1. Based on forum behaviour

Learners tend to express their sentiments towards a MOOC through the course forum, which offers a good opportunity to verify whether forum discussions are correlated with achievement. We select five state-of-the-art studies that explore this aspect. These studies are summarised in Table 1. As can be noted, each study investigates different aspects of forum behaviour. Most evaluate the impact of such factors by looking at student dropout. Apart from [10], all studies identify forum behaviour variables that influence learners' achievement.

#### 3.2.2. Based on activities on the MOOC platform

As mentioned previously, MOOC platforms unobtrusively collect a large amount of data from learners' interactions with the platform. A large number of works try to exploit this data to infer learners' achievement from it. We select 12 papers that are representative of this group. They are summarised in Table 2. Each work looks at a different subset of variables extracted from activities on the MOOC platform, and the way to measure achievement is also quite different from work to work. Due to the widely different data, methods and experimental setups, it is very difficult to compare works. That said, most works find variables that can be used for early detection of at-risk students, which is useful for MOOCs.

#### 3.2.3. Based on activities on the MOOC platform and background info

Every learner is different, and that's even truer for MOOCs. As such, taking into account learners' background may provide some insight into achievement. However, [11] showed that looking at learners' demographics alone is not sufficient, as this provides very limited explanatory and predictive information about achievement in MOOC. Instead, learners' background should be coupled with the activities learners engage in on the platform; in some ways, background is used to augment the behaviour of learners on the platform. We select two papers that took this approach; they are summarised in Table 3. While it appears that demographics do not show any clear correlations with achievement, other background information (such as prior competence in the subject matter) has been shown to help explain and predict achievement in MOOCs.

#### 3.2.4. Based on course and platform characteristics

Finally, some studies look at whether course and platform characteristics influence achievement (that is, elements that are independent of the course content). We review four such papers; they are summarised in Table 4. As can be seen, course characteristics such as due date, difficulty, and workload are correlated with achievement. Similarly, platform characteristics are also correlated with achievement. Gamification approaches also hold potential to influence achievement in MOOCs.

#### 3.2.5. Summary

In this section, we reviewed various methods to model, predict, and influence learner achievement, which is a crucial challenge in improving MOOCs. The exact data sources used for this purpose are unlimited, as was demonstrated by the different state-of-the-art methods described, each employing different data.

It is important to note that for most of the studies presented, links established are correlational in nature (as opposed to causal). What this means is that the process that leads to learner achievement (and its underlying explanation) in MOOC is still poorly understood.

Another point to note is that most works rely on a very limited set of input variables to build their models, especially when compared with the amount of data available in MOOCs. Apart from a few exceptions, they also test their models on very few MOOC courses. Richer models that take advantage of more data might hold the potential to better predict and influence achievement, although such models would generally need to be tested on more MOOC courses to avoid overfitting.

Finally, as mentioned in [6], it should be noted that measuring achievement through course completion or final grades (the two primary methods used currently) is probably too restrictive a method. Learners may achieve their goals without passing the course or completing graded exercises, which indicates that there could be value in designing new achievement metrics to obtain a richer understanding.

## 3.3. Modelling Learners' Knowledge

While better understanding learner engagement and achievement is important, it is far from sufficient to determine whether a MOOC is successful in its primary goal: delivering new skills and knowledge to learners. Indeed, a learner may well be engaged in and complete a MOOC without having learned much, if anything. As such, modelling learners' knowledge and skills is a crucial aspect to determine a MOOC's success. However, this is not trivial (and arguably much more difficult than the previous two problems). Some research has taken place towards this goal. Once more, we categorise it according to the type of data used for modelling learners' knowledge.

### 3.3.1. Based on assessed activities alone

An obvious strategy to measure and model learners' knowledge is to look at their assessed activities (homework, quizzes, exams, etc.) Most courses require learners to complete such activities in order to complete the course, so it is a natural strategy. We review three papers that use this strategy. All three of them evaluate their model by building a predictive model based on performance on homework problems, and testing it on exam problems.

In [12], knowledge is measured using discrete knowledge components (KCs) defined by a subject matter expert. Each assessed problem is a KC, and a Bayesian Knowledge Tracing based method is used to model learner knowledge. They show that they can predict with good accuracy whether a learner will successfully answer an exam problem (i.e., whether she acquired that knowledge during the course). This method requires expert input, which may not always be possible.

In [13], modelling domain-specific knowledge (namely, programming) is investigated. They use a variant Additive Factors Model to automatically extract a domain model by exploiting the structure in the Java programming language and then model student knowledge. They show that their models outperform a state-of-the-art model that does not assume learning is taking place. Compared to the previous method, this method is fully automated, but it can only be applied to the programming domain.

In [14], performance on homework is used as a proxy to measure knowledge. Models based on item response theory are designed to overcome the issue that data is missing for a large number of homework problems and that multiple attempts are allowed for each problem. Evaluation is done the same way as for [12], and their results show that their models offer improved estimates when compared to purely correlational models.

### 3.3.2. Based on activities on the MOOC platform as well as survey data

Another approach is not only to look at assessed activities, but to consider all activities learners engage in on the MOOC platform. Data obtained from pre- and post-course surveys can also be used. We review five studies that use this approach.

In [15], knowledge is deemed to be related to the scores on assessed components due to the way these assessments were designed. The authors then look at the number of activities learners engage in as well as learners' feedback collected from a post-course survey. Regression analysis and classification are then used to find variables that correlate with knowledge. A higher number of online activities is linked with more knowledge gain. Learners' perceived usefulness (i.e. whether the course appears valuable to the learner) is also linked with better learning outcomes.

Similarly, in [16], assessment grades are used as a proxy to knowledge. Through exploratory data analysis, it is found that students learning through performing tasks achieve better learning outcomes than those only viewing informational assets (e.g. videos).

[17] proposes to measure knowledge using the Precise Effectiveness Strategy, a methodology that uses metrics to calculate the effectiveness of learners when interacting with MOOC materials. These metrics must first be defined by experts before they can be computed automatically, based on learners' activities. At this point, no validation experiment has been carried out to verify whether this approach actually correlates well with knowledge.

In [18], knowledge was assessed using a purposely-built assessment tool that was administered both before and after the MOOC. The authors then tried to identify factors that can predict knowledge gains. Using an OLS regression model, they found that all predictive factors relate to effort in the course (as measured by the number of activities), prior courses taken, and baseline knowledge.

In [19], the goal was to identify which learning activities are the most useful for learning. They measured knowledge in multiple ways: assessment scores, improvement in skills (using Item Response Theory), and improvements in conceptual understanding (using a pre- and post-test). Wide differences in demographics and initial skills of learners prevented authors from drawing any conclusions about time spent on resources and knowledge gain.

#### 3.3.3. Summary

Compared to measuring behaviour and achievement, measuring knowledge has received less attention so far, probably due to the difficulty of the task. Most works can be considered preliminary and need to be further expanded.

To do so, collaboration with experts from the educational domain will be necessary, as pure AI and DM is unlikely to lead to good outcomes for this task. For instance, while DM can look at assessment scores to infer knowledge gained, nothing says that these assessments were designed in a way that could measure knowledge gain.

### 3.4. Summary

In this section, we reviewed the state-of-the-art work for better understanding MOOC learners, looking at it from three perspectives: 1) modelling engagement and learning behaviours, 2) modelling, predicting and influencing learners' achievement, and 3) modelling learners' knowledge. To conclude this section, we provide a visual overview of the AI and DM tools used for each aspect, based on the taxonomy presented in Section 2. This is summarised in Table 5.

## 4. Course Contents

Course contents are the primary learning resource for learners. By their virtue of being hosted online, MOOC contents can benefit from AI and DM. In this section, we describe state-of-the-art AI and DM tools applied to course contents, looking at the different types of contents present in MOOCs.

### 4.1. Videos

Videos are undoubtedly the most prominent learning content available to MOOC learners. Most courses rely on videos to deliver content at scale. We identify two main areas in which AI and DM have been used to improve MOOC videos: video pattern analysis, and improving video navigation and watching.

#### 4.1.1. DM for video pattern analysis

With millions of learners watching thousands of videos, there is a wealth of raw data available about video watching activities. Surely, interesting patterns can be discovered from this data.

In [20], authors study video production features and correlate them with measures of engagement (how long learners watch a given video, and whether they attempt a post-video exercise). Through the use of simple statistical tools, they make a number of useful observations that can help instructors create more impactful MOOC videos. These observations mention that shorter videos lead to more engagement, and specify production styles that can lead to more engagement. This can help instructors create MOOC videos that maximise learner engagement.

[21] also finds a strong correlation between video length and engagement: the longer the video, the higher the in-video dropout rate (i.e. a learner does not finish watching the video). They also find a difference between videos watched for the first time compared to re-watches, which lead to higher in-video dropout rates. Using binning and kernel-based smoothing, they then show second-by-second plots of video interaction (play, pause and skip events) peaks, which can show the typical behaviour of learners within a video. They manually categorise each peak in 5 categories that explain the underlying cause of the peak. Their analysis enables them to better understand how students interact and learn, which could prove useful to offer better video interfaces.

Course instructors may also get value from visualising video viewing data for their own MOOC. This may allow them to understand which videos are most useful, which parts of videos appear to create confusion, and more, eventually enabling them to improve their course videos. [22] provides a visual analytical system for this purpose. Their system simply computes a number of statistics and then presents them in a graphical and interactive manner.

#### 4.1.2. AI to facilitate video navigation and watching

Understanding video patterns is certainly useful but it does not directly provide value to learners. One way to achieve this is by adding AI tools to facilitate video navigation and watching.

[23] aims to facilitate non-linear video navigation by augmenting the video interface with different tools: a customised dynamic time-aware word-cloud and a 2-D timeline (to provide an overview of the concepts discussed in the video and allow quick navigation to these concepts), video pages (to enable visually searching for information by using automatically extracted visual slides) and a video summarisation method (to provide a summary of the video).

[24] automatically identifies and links related topics in MOOC videos. It then creates an interface that enables learners to navigate to topics of interest. Their approach is based on fuzzy formal concept analysis and semantic technologies.

### 4.2. Supporting Materials

MOOC videos are often complemented with supporting materials such as lecture notes and reading materials. AI can help organise and even generate such materials.

To enable more flexible navigation and easier retrieval, AI can be used to label and index materials. This is the approach presented in [25], in which Collaborative Semantic Filtering technologies are used to enrich MOOC materials with semantics, thus enabling collaborative labelling and indexing.

Supporting materials can also be automatically generated, thus saving instructor time. In [26], the authors present a method to automatically generate an interactive presentation by using course video slides. The method is based on a topic structure analysis model and enables the creation of topic summaries based on semantics.

### 4.3. Formal and Informal Assessment

After consuming the informational course assets, MOOC students are generally invited to apply the new knowledge through hands-on exercises. These can be assessed or not, and can be formal or informal, depending on whether a given exercise makes up a part of the course grade or not. We will focus our discussion on materials that are assessed (i.e. there is a mechanism to verify the correctness of the solution and/or feedback is offered to the student along the way or at the end) since this is an obvious area in which AI and DM can contribute – and indeed much research has already taken place.

Current ways in which assessment is carried out in most MOOC platforms include 1) auto-graded content with predefined answers (e.g. multiple-choice questions, numerical questions, short-answers with limited solutions, etc.); 2) auto-graded content with open-ended answers (e.g. essays, derivations, programming); and 3) peer-graded content with open-ended or non-trivial answers. Current AI and DM research mainly focuses on solutions for the last 2 (since the first one is a trivial problem for which adequate solutions already exist), as well as providing tools to instructors to help with assessment.

#### 4.3.1. Auto-graded content with non-trivial answers

For this kind of assessment, research is always targeted at domain-specific problems since each domain has specific requirements. On top of checking for answer correctness, feedback may be offered, either along the way (as hints to get to the correct solution) or at the end (to suggest improvements).

Programming is by far the domain that has received the most attention. While testing the validity of a code program is a trivial problem that can be solved through the use of test cases, objectively evaluating the quality of a program is much more difficult. [27] does this by considering a program to be constituted of multiple code fragments and measuring the distance between fragments of different students through appropriate distance metrics. By relying on previously graded programs, a quality score is assigned (i.e., programs that have a small neighbouring distance should be of similar quality).

However, most works related to programming focus on how to automatically provide feedback to learners. [28] tries to predict how an instructor would encourage a learner to progress towards the correct solution to automatically generate hints. They use Desirable Path algorithms to model the best paths that can lead to the correct solution and steer learners in that direction. In [29], authors develop a technique that uses an error model describing the potential corrections and constraint-based synthesis to compute minimal corrections to student's incorrect solutions, enabling them to automatically provide feedback to learners. Their results show that relatively simple error models can correct a good number of incorrect solutions. Finally, in [30] authors use a dynamic analysis based approach to test whether a student's program matches a teacher's specification. After manually identifying specifications and their associated feedback, the algorithm matches a program with a specification and automatically generates the relevant feedback.

Embedded systems have also received some attention in this area. In [31], authors use constrained parametrised tests based on signal temporal logic that capture symptoms pointing to success or failure. They can then automatically generate the relevant feedback and grade the exercise. [32] investigates a different problem: how to automate exercise creation, solution generation and grading. They do so by generalising existing exercises from textbooks into templates that capture the common structure. They then adapt previously developed techniques for verification and synthesis to generate new problems and their solutions.

A few works have looked at automatically grading mathematics problems. In [33], authors investigate how to automatically grade open response mathematical questions. They leverage the high number of solutions created by learners to evaluate the correctness of their solutions, assign partial-credit scores, and provide feedback to each learner. Their method is based on clustering solutions based on extracted numerical features. [34] instead focuses on automatically grading and offering feedback for derivations. However, at a high level, their approach is similar to [33]: they leverage the high number of solutions to provide real-time feedback for derivations.

#### 4.3.2. Peer grading and feedback

When automated grading is not feasible, peer grading is usually the chosen alternative. In peer grading, a learner’s assessment is reviewed and graded by other learners. Usually, on top of the grade, detailed feedback is also offered. A common challenge in peer grading is how to do grade aggregation so that the outcome is as accurate and fair as possible. A number of works have tackled this task.

In [35], authors use a trust graph to automatically combine grades from peers and tutors. The trust model is used to compute how much weight to give to each grader when coming up with the final grade. [36] develop methods to improve both cardinal (absolute judgment) and ordinal (relative judgment) peer grading. On one hand, they extend existing probabilistic graphical models to improve cardinal grading. On the other hand, they use cardinal prediction priors to augment ordinal models. [37] seeks to lower the grading burden on peers while maintaining quality. It does so by using a machine learning algorithm to automatically grade an entry. Peers then identify key features of the answer using a rubric, and other peers verify whether these labels seem accurate. Depending on the algorithm’s confidence and peer agreement, a different number of peer graders is assigned for each entry. Finally, [38] develops a series of probabilistic models that estimate and correct graders' biases and reliabilities. All of these methods have been shown to improve peer grading accuracy. However, since they do not compare themselves again each other, it is difficult to judge which one is the best.

Another branch of work related to peer grading is the development of tools that can support the peer grading process. In [39], automated tools guide the peer assessment process for writing tasks. Their system automatically scaffolds a paper based on predefined rubrics that target both structural and relational criteria, highlighting relevant content and indicators for graders. At this point, no experiments have validated whether this tool improves the accuracy of peer grading or not.

#### 4.3.3. Instructor support

Due to the very high learners-to-instructor ratios in MOOCs, it is not realistic to expect instructors to manually grade assessments. However, there are reasons why instructors might want to take a look at assessments, such as to allow them to better understand the gaps in learners’ knowledge, or to provide targeted feedback at scale. There are also times when instructors must provide some input. For example, as discussed previously, some algorithms for automatic assessment of open-ended problems require instructors to provide a set of graded solutions. Without tools, these tasks become very difficult and inefficient due to the large number of entries. AI and DM can be used to support instructors in this task.

Clustering can be a powerful tool to quickly see what learners understood, to identify patterns in solutions, and to provide feedback at scale. In [40] and [41], authors develop a clustering-based visualisation tool that enables one to view functionally similar programming solutions. Their algorithm is unsupervised, meaning it does not require any human input to operate. [42] uses a data-driven probabilistic approach  to index 'code phrases' from programming assignments and semi-automatically identifies equivalence classes across these phrases. Instructors are required to label a set of abstract syntax tree (AST) subtrees that are semantically meaningful before they can query all entries to extract similar subtrees. As such, this is not a true clustering tool since it is supervised. In [43], the authors generate an AST for each solution to a programming problem, and calculate the tree edit distance between all pairs of ASTs, using the dynamic programming edit distance algorithm from another work. Based on these computed edit distances, clusters of syntactically similar solutions are formed. However, their approach is computationally expensive (O^2).

In [44], authors cluster short-answer problems to enable the instructor to grade each cluster at once, instead of going through all entries. Their clustering algorithm computes a learned distance metric over pairs of answers, based on different word characteristics, and then uses a version of the k-medoids algorithm to generate the clusters. Through this approach, grading is made much faster (but still requires human input).

### 4.4. Adaptive Materials and Sequences

As mentioned previously, learner characteristics vary a lot in MOOCs. Different learners may have widely different backgrounds, and different reasons for taking MOOCs. As such, it seems unlikely that a one-size-fits-all approach can cater to all learners. However, due to technical and resources limitations, this is currently the case in most MOOCs: all learners go through the same materials in the same sequence, and are assessed the same way. The case can be made that providing adaptive MOOCs (i.e. MOOCs for which the materials and presentation sequences are personalised for each learner) holds real potential in increasing the quality and retention of MOOCs. This issue has attracted research attention in the last few years, and AI and DM can clearly contribute to this task. All works aim to build a user model that can then be used to adapt the learning path.

One team has looked into strategies to provide adaptive video sequences. In [45], authors use an EEG headset to measure learner's attention while they watch videos. They then adaptively suggest a video review sequence that could maximise learning. Their results showed improved recall as well as learning gains as a result of the proposed review system.

Another team investigated the automatic generation of assessment objects and remedial works. By modelling learners' knowledge gaps (through the automatic generation of assessment questions based on subject ontologies) as well as preferences, their system can then generate sequences of remedial work that are adapted to each learner. Education experts subjectively evaluated their approach and found it to be valuable [46].

In [47], authors propose a reinforcement learning based algorithm to analyse every learner's profile. Using both implicit and explicit feedback, they infer learners' needs and capabilities and then recommend an appropriate learning sequence. They validated their results by asking learners to rate the usefulness of recommended learning objects, showing that ratings were higher after recommendation compared to before.

In [48], authors build their learner model by using machine learning techniques on behavioural data (i.e. activities on the MOOC platform). They then provide personalised content based on this model. Preliminary results indicate that learners subjectively prefer the adaptive platform over a traditional one, and that they statistically view more content.

In [49], instead of trying to model each learner individually, authors classify them in one of five distinct learning strategies through an assessment tool. They then provide an adaptive sequence for each strategy. The sequences must be provided by an instructor. No results are reported about the effectiveness of this system.

[50] relies on open repositories of learning objects to build adaptive sequences. Collaborative filtering is used to model learner preferences. Based on these preferences, repositories are queried by exploiting the metadata attached to each learning object. No results are reported.

Finally, [51] replaces the traditional MOOC environment with an adaptive serious game. A learner-centred numerical model that considers skills (measured through instructor-defined competency maps), explicit user preferences, and learner behaviour is used to provide an adaptive learning path. No results are reported.

Providing adaptive learning paths in online education systems is not a new field; in fact, much research has already taken place in various environments. MOOCs have an advantage over other systems due to the vast amount of data available. The real challenge is to find efficient and effective ways to exploit this seemingly unstructured data.

### 4.5. Content Authoring and Creation

As mentioned in the introduction, preparing the content for a MOOC can be a daunting and time-consuming task. AI and DM can help make this process more efficient, in different ways.

First, AI can be used to automatically create contents. We described an example of this in section 4.3.1: [32] can automatically create new exercises based on templates.

Second, AI can be used to find and reuse existing contents. There already exist a large number of open learning resources on the Internet, which represents a great opportunity to tap into when authoring MOOCs. However, finding such resources is often difficult, and integration can prove to be a challenge as well. The use of taxonomies, ontologies, and semantic technologies can help. [52] presents the main aspects needed to support the discovery, accessibility, visibility, and reuse of open resources in MOOCs. Their framework, based on semantic web technologies, apply the principles of Linked Data for these tasks.

Finally, although this is not a pure AI approach, the concept of crowdsourcing can help. In crowdsourcing, content creation is ‘outsourced’ to many people, such as other instructors or the learners themselves. AI can come into play when it comes to determining when someone can contribute, and whether that contribution is of quality, for instance. [53] approached the first issue. They developed a crowdsourcing system to allow learners to generate hints. Their system identifies learners that first get an exercise wrong, and then manage to get the correct solution. It then invites the learner to contribute a hint that helped them to get to the correct answer.

Expert evaluation revealed that the contributed hints were generally of high quality, although no automatic quality verification is performed.

### 4.6. Summary

In this section, we reviewed the state-of-the-art AI and DM work for MOOC contents. We looked at course videos, supporting materials, formal and informal assessments, adaptive materials and sequences, and content authoring and creation. To conclude this section, we provide a visual overview of the AI and DM tools used for each aspect, based on the taxonomy presented in Section 2. This is summarised in Table 5.

## 5. Community Building

The large community is what differentiates MOOCs from other online learning systems. This is a valuable resource that needs to be harnessed, as it can improve the experience and the learning outcomes of MOOC learners. In this section, we review AI and DM tools for the different types of community building components that can be found in MOOCs.

### 5.1. Asynchronous Tools

Forums are the main asynchronous tool to provide interactions in the learning community and are used by most MOOC platform (in fact, it's the only such tool we identified in current MOOC platforms). AI and DM research for MOOC forums has focused on three main tasks: analysing forum involvement over time, understanding forum users, and improving forum welfare.

#### 5.1.1. Analysing forum involvement over time

Understanding the trends of how learners get involved in the course forum throughout the duration of a MOOC is an interesting issue. It is also a first step in taking action to encourage fruitful involvement. Unsurprisingly, research finds that involvement declines as the course progresses, in a similar fashion as learner dropout. In [54], authors split forum posts into the common MOOC forum subcategories (e.g. study groups, lectures, assignments, etc.) and then carry out a statistical analysis of the number of posts in each subcategory as a function of normalised course length. They show an exponential decay of posts across all subcategories. In [55], authors took a similar approach, although with slightly different subcategories. They also find a sharp decline across all subcategories as the course progresses. They also use linear regression to identify factors that help to explain this decline. Interestingly, they find that active participation of the instructors is associated with an increase in discussion volume but does not reduce the participation decline rate.

#### 5.1.2. Understanding forum users

Understanding trends is a good first step but is not sufficient to act at an individual level. To do so, we must understand forum users themselves. In [56], authors investigate the contributions of MOOC 'superposters', i.e. the most active users in a course forum. Through statistical analysis, they demonstrate that high superposter activities correlate to high overall forum participation and health, and that superposter entries tend to be of high quality. As such, encouraging such posting behaviour would be valuable. In [57], authors use content analysis based on natural sociological inquiry to understand whether forum interactions are mostly about course content or not. Their results showed that most threads were indeed related to course contents and as such had value for the learning experience.

#### 5.1.3. Improving forum welfare

Finally, after better understanding what goes on in forums, it is possible to act to improve forum welfare. Most of the AI and DM works target this problem, from different perspectives.

The most popular strategy to improve forum welfare is through identifying and suggesting relevant forum contents to learners. The high volume of forum posts often prevents learners from keeping up-to-date, so mechanisms are needed to ensure they can view the most relevant contents. In [58], authors use linguistic modelling to identify content-related threads. They then do classification on manually labelled threads to demonstrate the effectiveness of their approach. In [54], authors propose a classification scheme of forum posts using non-text features that is language independent and that is reasonably accurate. It can be used to ensure that the labels selected by learners are accurate, thus improving navigation. [59] uses an unsupervised clustering approach to group similar posts. The clusters thus identified do not fit under the traditional forum subcategories but are instead related to the content (e.g. a cluster where learners agree with one another, one that contains questions, one that emphasises important contents, etc.), which can be useful to decide which threads need more contributions or should be viewed in priority. In [55], authors build a generative model that enables them to efficiently classify threads as well as assign them a relevance ranking. This can facilitate user navigation by only showing learners the most relevant threads. [60] specifically targets question answering in forums. They develop a constrained question recommendation algorithm based on a context-aware matrix factorisation model that predicts students' preferences over questions, as well as a model to optimise community benefit under multiple constraints. Their results outperform other baseline approaches for question recommendation.

Forum welfare can also be improved through directly providing help to forum users. In [61], authors develop an analytics system that clusters forum threads to help instructors identify which students are contributing, struggling, or are distracted, allowing them to target their interventions.

Finally, one approach to improve forum welfare is through incentive mechanism. In [62], authors added a reputation system to the MOOC forum in a way that is similar to that of e.g. Stack Overflow[5]: learners can upvote or downvote a post, and can select a best answer. The reputation score is then calculated by taking these features into account. Their results showed that the reputation system leads to more and faster responses, although it did not help to improve grades, retention or sense of community. In [6], authors deployed four different badge systems to encourage desirable behaviours in MOOC forums (e.g. reading, posting, answering, voting). They showed that the more prominent badges are, the more engagement there is in the forum.

---

[5] http://stackoverflow.com/

## 5.2. Synchronous Tools

Synchronous tools, although less popular than asynchronous tools, can also be found in some MOOC platforms. We identify two main types of synchronous tools: chat rooms, and small group discussions.

### 5.2.1. Chat rooms

Chat rooms can be seen as an 'uncontrolled' type of synchronous tools, in the sense that they allow any learners to interact with one another, as long as they are online at the same time. There are usually no restrictions on chat rooms.

In [63], authors make a chat room available throughout different course offerings, enabling learners from different cohorts to interact with one another. Through statistical analysis, they demonstrate that participants from previous offerings often step in to help current learners, which shows that communities of practice can organically emerge through chat rooms. [64] used a chat room to complement the forum, but their conclusion was different. Their statistical analysis demonstrated that there is no significant correlation between chat use and variables such as grades, retention, forum participation, or students' sense of community.

### 5.2.2. Small group discussions

Small group discussions, on the other hand, can be considered a 'controlled' type of synchronous tools, since there are usually parameters controlling how such groups can come together.

In [65] and [66], authors develop a system for small group video discussions. The groups are formed based on specified instructor preferences, such as gender balance and geographic distribution. Statistical analysis revealed that learners spent more time using the tool than the course requirement, showing the value they attach to it. Furthermore, learners in groups that comprise of geographically distributed learners get higher assessment scores. Other similar systems have been proposed (e.g. [67]) but they do not report results.

## 5.3. Group Work

Group work is based on the concept of project-based learning and has been confirmed to be an effective educational approach [need ref]. In MOOCs, AI and DM can help in automatically grouping students in an effective manner.

In [68], authors develop constraints-based team formation principles and algorithms so that productive, creative, or learning teams can be automatically formed. They rely on data about the project and learner knowledge, personality and preferences. Experts validated their approach for productive and learning teams but not for creative teams. [69] also introduces a methodology for dynamic team formation in MOOCs, through the use of organisational team theory, social network analysis, and machine learning. The overarching team formation principles aim to connect people that possess different skills and that do not exhibit historically strong relations with one another. They do not report results yet.

### 5.4. Peer Support

Finally, peer support (in the form of mentoring and tutoring) can play a role in building a sense of community in MOOCs. This kind of support can happen through both synchronous and asynchronous tools, and can be done through tools that are fully dedicated for this purpose or not. For instance, peer support can happen through forums, even though such tools have other objectives as well. AI and DM can play a role in optimising the peer support process.

While peer support can happen between learners themselves, or between a learner and an intelligent agent, to the best of our knowledge such research has not been applied to MOOCs yet. Instead, research focuses on support from instructor to learner. Instructors have limited time, so tools are needed to ease the process. In [70], authors developed a code sharing tool that allows learners to easily share code with instructors as well as the issues they are experiencing. Instructors can then quickly visualise the issues and offer personalised feedback to the learner. Through this streamlined approach, a limited number of instructors were easily able to handle all help requests, showing its effectiveness. On the other hand, [71] challenged altogether the assumption that instructor involvement actually matters in MOOCs. Through A/B testing and statistical analysis, they demonstrated that instructor participation has no statistically significant impact on dropout rates, participation, or satisfaction. This provides an alternative view on whether it is important to allow instructors to interact with learners.

### 5.5. Summary

In this section, we reviewed the state-of-the-art work for building thriving communities in MOOCs, looking at it based on the types of community building components available in different MOOCs. We reviewed asynchronous and synchronous tools, group work tools, and peer support tools. To conclude this section, we provide a visual overview of the AI and DM techniques used for each aspect, based on the taxonomy presented in Section 2. This is summarised in Table 5.

## 6. Platform

Finally, we turn our attention to AI and DM research applied to the broad MOOC platform framework. We identify two elements to discuss: 1) course recommendation and search, and 2) student authentication and cheating detection.

### 6.1. Course Recommendation and Search

Given the high and ever-growing number of MOOC courses and platforms, learners have an abundance of choice when it comes to learning opportunities. However, this can also be seen as a challenge – at times, it can be overwhelming looking for courses, and choosing a platform. Oftentimes, learners know what they want to learn about and are looking for courses that would allow them to fulfil their needs. Other times, learners might want to get recommendations for courses to take that they might find interesting and that would complement their current knowledge.

#### 6.1.1. Searching for MOOCs

All major MOOC platforms offer a basic search engine that enables learners to look for courses. Also, many MOOC aggregator websites currently exist, allowing searching for courses across multiple platforms. In general, MOOC aggregator websites provide more advanced searching functionalities.

There exists some research about searching for MOOCs across platforms. In [72], authors build a vertical search engine that crawls, parses, and indexes MOOCs from many sources. In [73], authors also develop a vertical search engine. They improve on the Term Frequency-Inverse Document Frequency algorithm by also considering the properties of the query terms: words are assigned a weight based on their importance within the query. [74] takes a different approach. They use semantic technologies to publish linked data about MOOCs and not only enable searching for courses, but also allow for direct comparisons between similar courses.

#### 6.1.2. MOOC recommendation

Again, major MOOC platforms provide recommendations for new courses to take. Although they do not disclose their algorithms, from personal experience it appears to be based on a learner's past history of courses taken on that platform as well as the learner's profile. A small body of research looking into course recommendation is available. In [75], the authors investigate how to recommend MOOC courses automatically. They use probabilistic topic models to infer course content from college and MOOC syllabi. They then use learners' grade records from college to match them with MOOC courses in order to offer them remedial courses. Their preliminary results demonstrate that the recommended MOOCs match well their equivalent college courses in terms of content. In [76], authors explore the concept of social recommendation, i.e. learners recommending courses to others through their social network. Through the use of correlation and regression analysis, they identify which MOOC attributes are the most valuable for social media sharing. They identify six such attributes and demonstrate that these can vary depending on the social media tool used for sharing.

### 6.2. Student Authentication and Cheating Detection

If MOOC providers want to offer formal recognition, then student authentication and cheating detection are both necessary. Major platforms already have authentication tools in place. For instance, Coursera and edX both use image recognition (i.e. comparing a webcam photo with a previously submitted government ID) to authenticate students. Coursera also uses typing pattern as a second layer of authentication. Some researchers have proposed alternative approaches for authentication. In [77], authors propose a global user authentication model that can include a wide range of authentication modalities (e.g. Turing test, biometrics, double verification, continuous user authentication based on MOOC activities, and more). Their model can select any number of modalities and integrate them in a seamless framework. No results are reported.

It is unclear whether major MOOC platforms also use cheating detection tools at this point. So far, as far as we know, no research has directly applied cheating detection tools to MOOCs, although a body of work does exist for cheating detection in online learning systems and for offline learning.

### 6.3. Summary

In this section, we reviewed the state-of-the-art work for improving MOOC platforms. We looked at course recommendation and search, and student authentication and cheating detection. To conclude this section, we provide a visual overview of the AI and DM techniques used for each aspect, based on the taxonomy presented in Section 2. This is summarised in Table 5.

## 7. Key Trends and Future Research Directions

A lot of work remains to be done before MOOCs can achieve their true potential, and both AI and data mining still have an important role to play. We now list a list of key trends and areas where new research and development could significantly improve MOOCs.

### 7.1. Redefining what 'Open' means in MOOCs

Current MOOCs can be considered open from the perspective outlined in the introduction – they are freely accessible to all learners. However, we would like to highlight other dimensions of openness that would be highly valuable as well.

We already made the point earlier that creating MOOCs is a resource-intensive process for instructors. In that sense, MOOCs are not fully open to all instructors, since they must have the resources before they can embark on the MOOC journey. When reviewing the current state-of-the-art work, we already outlined many areas where AI and DM ease the process for instructors, from content creation to course management. However, more needs to be done, as the majority of this work is preliminary and not easily scalable. We believe that the biggest contribution in this area would come from improving content reusability and interoperability. There already exist a lot of open materials available online (and more is created every day) but finding it and integrating it in MOOCs is largely impractical currently. Standardising knowledge representations specifically for MOOC contents would be a good first step in this direction.

Openness for researchers is also a critical issue if we want to improve MOOCs. Specifically, access to data is essential. Most works use data from a very small subset of MOOCs since this is the only data they have access to. Each dataset uses different data sources, and the underlying characteristics of the courses covered by each dataset vary widely from dataset to dataset. What this means is that comparison between studies is very difficult, and the validity and generalisability of results is unclear since the data used to obtain them might not be representative. Open benchmark datasets, as is common in many other fields, could certainly help overcome these issues.

On top of the technical challenges associated with the tasks described above, it also requires concerted efforts by MOOC providers and institutions, as well as policy changes. We are well aware of legal issues that currently hinder these changes from taking place (copyright issues, privacy laws, etc.) but feel it is necessary to discuss how to overcome these barriers creatively so that MOOCs can reach their full potential.

### 7.2. A more multidisciplinary approach to MOOC research

There already exists a large body of work on education for online paradigms, and an even larger body of research on education in general. So far, MOOC AI and DM research has not exploited this well. Few tools that have proven effective in online learning systems (such as intelligent agents) have been tested or integrated in MOOCs. Research from learning analytics has not been leveraged, MOOC researchers mostly trying to develop their own analytics as opposed to starting from well-established research. We could list more examples. Operating in silos will not allow MOOCs to reach their full potential.

We believe that AI and DM for MOOCs should be used in conjunction with existing research from other related fields. For example, AI and DM could be leveraged to test educational theories at scale, and look for causal relationships instead of correlational ones. MOOCs could also be used to re-test research from smaller-scale online education systems, to see whether the results scale well. MOOCs have an asset that most education research struggles to acquire: vast amounts of data. Researchers need to make better use of this.

### 7.3. Complementing AI and DM with humans when necessary

AI has improved significantly recently but there are still a number of challenges that it is unable to solve. For such challenges, combining the power of AI with humans (a concept known as crowdsourcing) could prove game-changing.

One important problem in MOOC research is the fact that most data is unlabelled. This makes it much harder to draw causation links in the data (as opposed to simple correlations). At the same time, supervised algorithms must thus rely on the small amount of labelled data (either provided by student surveys or researchers themselves – both approaches are expensive). Examples of data labelling through crowdsourcing include allowing students to rate different portions of a video to assist curriculum redesign, gathering insights about navigation behaviour through short pop-up questions, collaborative materials annotation to label contents, and more.

Crowdsourcing could also be leveraged to reduce instructor workload. We described an example of this in section 4.5, in which learners were leveraged when creating hints. This approach could be expanded to other aspects of MOOCs such as exercise creation. Game theory (through incentive mechanisms), trust modelling frameworks, and rating mechanisms could be put in place to ensure the quality of the contributions.

### 7.4. From engagement to knowledge

As showcased in Section 3, the majority of MOOC research looking at better understanding learners is concerned with engagement, achievement, and how students navigate through MOOC platforms. While this information can be valuable for a number of reasons, it is not sufficient to ensure that MOOCs are in fact able to impart knowledge to students. A student can be engaged in a course without actually learning much (e.g. if the material is too simple). Automatically assessing the instructional quality of MOOCs and identifying the criteria that lead to quality courses could be valuable. Currently, such assessments are done manually, which is not suitable at MOOC scales. More targeted questions could also be investigated. For example, when it comes to the usefulness of hints and tips to impart knowledge, current results are conflicting, so new research is needed to shed some light on this question.

We saw in section 3.3 that although some research has taken place in modelling learner knowledge, much of it is preliminary and domain-specific. More needs to be done so that methods can be deployed easily across domains. The current research also omits important elements related to learner knowledge. Indeed, all work focuses on knowledge gain as it pertains to the course materials, usually measured through purposely-built assessment tools. However, no research attempts to identify new knowledge and skills developed at a higher level (i.e. beyond technical competency in the subject matter, e.g. $21^{st}$ century skills and informal learning events) as a result of learners engaging in MOOCs. It could also prove valuable to look into which variables can predict that a learner will apply what has been learned as opposed to that knowledge remaining theoretical in nature.

### 7.5. More personalisation

Personalisation is important so that every learner has a fruitful experience in MOOCs. Previously, we made the case for providing adaptive contents to learners. Different learners do not learn the same way, so it makes sense to provide each one with a different learning experience. This is probably the most obvious form of personalisation in MOOCs, and more research should be done to improve it. As presented in section 4.4, current research is somewhat limited and generally requires significant human input to achieve adaptability. More automated approaches that are generalizable would provide high value to MOOCs.

Other forms of personalisation could be valuable. For instance, using intelligent agents has the potential to help in all aspects of MOOC. By their nature, such agents can mitigate the lack of instructor resources by taking some of their responsibilities. For example, affective agents could help students stay engaged and motivated throughout the course by offering personalised support (both emotional and technical), teachable agents could enable students to gather knowledge through the learning-by-teaching framework, and curious agents could help students discover interesting course contents and forum questions by recommending useful contents that the student has not yet explored. Other intelligent agents could improve forum welfare by automatically answering questions (e.g. from other online forums or previous offerings of the course), and asking questions that steer the discussions towards high quality learning interactions and increased participation (e.g. from historical data from previous MOOCs). There already exists a large body of research on the development and application of intelligent agents for online education paradigms. It is likely that a good part of it could be repurposed and integrated into MOOCs without requiring considerable effort.

Finally, personalisation in the form of automatically linking learners with one another could be valuable. One example would be mixing learners with different skills to promote peer help and learning-by-teaching (e.g. partnering a learner that got the correct answer to an exercise to one that did not, or forming small groups on the fly in a way that puts together students with different backgrounds and abilities). Another example would be matching peer reviewers with answers from which they may learn something. Doing this type of personalisation would first require a better understanding of how learners perceive learning with one another. It may also require designing incentives to encourage such behaviours.

One note of caution: while adding personalisation features has some clear benefits, their potential negative impacts should also be evaluated. For example, if learners all view different learning contents and in different orders, how would it affect the community interactions? Such impacts have not been considered until now.

## 8. Conclusion

In this paper, we reviewed the state-of-the-art artificial intelligence and data mining research applied to MOOCs, emphasising how AI and DM tools and techniques can improve student engagement, learning outcomes, and our understanding of the MOOC ecosystem. We showed how AI and DM are seamlessly embedded in virtually every aspect of the MOOC ecosystem as we know it today. We then gave an overview of key trends and important future research in AI and DM for MOOCs so that they can reach their full potential.

That said, MOOCs are not a panacea; they will likely never replace formal education as we know it. However, they are certainly one part of the puzzle to making universal, lifelong education a reality. By enabling learners from around the world to access free, high-quality courses, we are giving people more opportunities, whether it's the chance to get an education that would otherwise be inaccessible to them, to gain new skills to further their career, or simply to satisfy their thirst for new knowledge. By connecting learners from all over the world and from all kinds of backgrounds, by enabling them to interact with one another and share ideas and knowledge, we are creating an enabling environment in which boundaries do not matter anymore, an environment in which everyone can do what humans are designed to do: learn. There is no better time to contribute to this exciting endeavour than now!

**Figure 1: The Current MOOC Ecosystem**

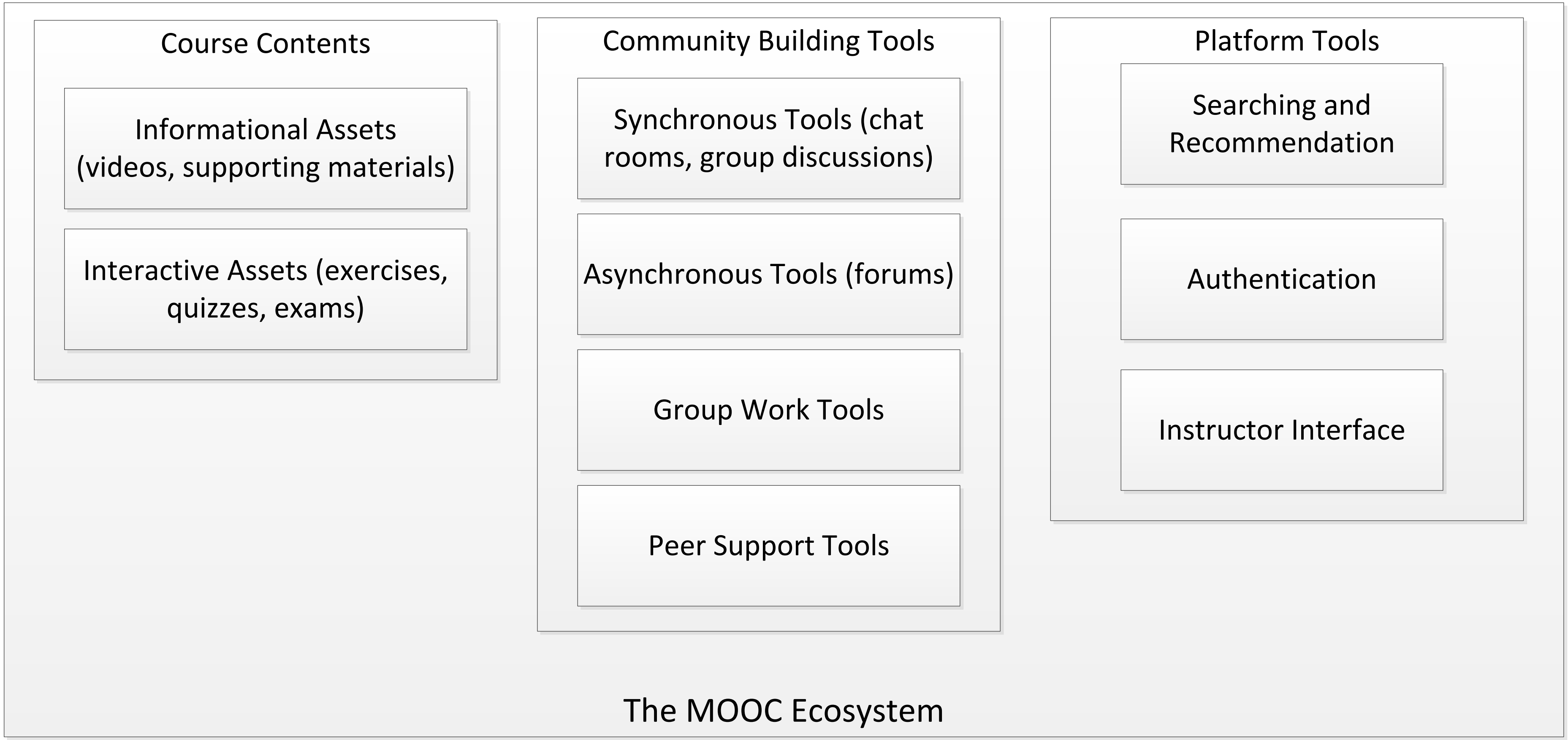

**Table 1: State-of-the-Art Research for Modelling, Predicting and influencing Learners' Achievement Based on Forum Behaviour**

| Ref | Model? | Predict? | Influence? | Individual or collective? | Features used | Method used | Achievement metric | Findings |
|---|---|---|---|---|---|---|---|---|
| [78] | X | X | | I | learner motivation and cognitive engagement (from the text of forum posts) | computational linguistic models; survival analysis | Dropout (per week) | Higher motivation is correlated with lower dropout.<br><br>Higher cognitive engagement is correlated with lower dropout. |
| [10] | X | X | | Both | students' opinions towards the course through forum posts | Sentiment analysis; Survival analysis | C: Dropout (per day)<br><br>I: Dropout (per week) | Use sentiment analysis with caution when looking at dropout rate since the data is noisy.<br><br>Results are not consistent across courses. |
| [79] | X | X | | I | Posting behavior (various measures), social network behaviour (various measures) | Survival analysis | Dropout (per week) | Social factors affect dropout along the way |
| [80] | X | X | | I | Soft partitioning of the forum social network; binary cohort variables; social network analysis measures | Mixed Membership Stochastic Blockmodel; survival analysis | Dropout (per week) | Participants in one particular emergent sub community are 10 times more likely to drop out.<br><br>Learners involved in the forum from Week 1 have lower dropout.<br><br>Learners with higher authority scores have lower dropout. |
| [81] | X | | | Both | textual data from forum posts | Sentiment analysis, statistical analysis that quantifies the correlation between | grades attained in course homework | Student sentiments are slightly (positively) correlated with quiz performance, and more strongly (negatively) correlated with homework assignments. |
| | | | | | | student sentiment and student performance | assignments, quizzes and examinations | |

**Table 2: State-of-the-Art Research for Modelling, Predicting and influencing Learners' Achievement Based on Activities on the MOOC Platform**

| Ref | Model? | Predict? | Influence? | Individual or collective? | Features used | Method used | Achievement metric | Findings |
|---|---|---|---|---|---|---|---|---|
| [82] | X | X | | I | e.g., number of lecture video views for current and previous weeks, finer-grained and temporal features | Random forests | Dropout (by week) | Finer-grained and temporal features boost early predictive performance, but only in certain subpopulations.<br><br>It is important to identify and analyse different subpopulations separately. |
| [83] | X | X | | I | 1) behavioral— constructed from user behavior such as posting in, viewing or voting on discussion forums, lecture views, and quiz completion; 2) linguistic—polarity and subjectivity values of forum-content calculated using Opinionfinder; 3) structural—constructed from forum-interaction; and 4) temporal—features from user activity over time. | probabilistic soft logic (PSL); survival analysis | 1) whether the learner earned a statement of accomplishment in the course, and 2) whether the learner survived the later part of the course | Monitoring learner activity in the middle phase of a course is most important for predicting whether the learner will complete it. |
| [84] | X | | | I | time-stamped logs of student activities such as lecture views, submission of assignments, participation in forums (e.g., threads views, posts, and up-votes logs), clickstream logs (logs for tracking user activity on the course Web site), page views, | Correlation | Grade | Delaying taking quizzes after they are posted is negatively correlated with achievement on quizzes<br><br>There exist significant contrasts in behavior between dropouts and non-dropouts |
| | | | | | lecture video interaction (e.g., video seek events), and geolocation information from IP addresses | | | |
| [85] | X | X | | I | Time spent on graded exercises, time spend on discussion, time spent on books, time spent on videos, time spent on ungraded exercises, attempts at exercises, demographics | Panel regression | Homework performance (per problem set) | Time spent on homework and labs is positively correlated with achievement.<br><br>Time spent on discussion, books and videos is a weak predictor of higher achievement.<br><br>Time spent on ungraded questions interspersed in lectures and number of attempts are strong predictors of higher achievement.<br><br>The results vary with demographics. |
| [86] | X | X | | I | Time series n-grams of lecture videos, discussion forum threads, and quiz attempts | time series interaction analysis, J48 classifier | Passing grade | Models are very accurate when used post-hoc but lack strong explanatory power.<br><br>Models trained on the first two offerings of a course are generalizable to a third offering with moderate accuracy.<br><br>Models are moderately accurate when applied to new real-world data by the third week of the course.<br><br>Day-by-day accuracy is characterised, which is important for building predictive modelling solutions. |
| [87] | X | X | | I | average quiz score for week 1, number of peer assessments | Logistic regression | Certificate (yes/no) + type | Assignment performance in Week 1 is a strong predictor of students' achievement. |
| | | | | | completed in Week 1, social network degree in week 1, whether or not a learner is an incoming UCI Undeclared major student | | | Social integration in the learning community in Week 1 is positively correlated with the achievement of Distinction certificates.<br><br>Students with external incentive are more likely to complete the course. |
| [88] | X | X | | I | 71 daily accesses, 25 three day accesses, 11 week accesses, and 3 month accesses, counts of the numbers of accesses on different days of the calendar week for discussion forums, quiz and lecture videos | n-grams, J48 decision trees | distinction (85% or higher) in final course grade (i.e. low achievers vs high achievers) | The model can accurately predict distinction.<br><br>Intra-course validity is limited.<br><br>The model has moderate accuracy as an early warning system |
| [89] | X | X | | I | user interactions with courseware as a “bag of interactions” using time spent in seconds on each course module | Latent Dirichlet Allocation (LDA) | certification | It is possible to predict whether or not a student will earn a certificate using only week 1 logs. |
| [90] | X | X | X | I | student engagement with video lectures and assignments, and performance on assignments by the end of each week | Transfer learning algorithms based on logistic regression | certification | LR-SIM is promising for early prediction.<br><br>The prediction models trained on a first offering of a course work well on a second offering. |
| [91] | X | X | | I | Interaction and persistence, based on number of viewed videos, downloaded lectures, and replayed quizzes and surveys | Statistics | Engagement degree (self-defined) | greatly reduced the latency time to analyse the huge amount of MOOCs’ generated data, allowing us to identify “at-risk” learners at different stages of learning operations through the MOOC platform in a reasonable time |
| [92] | X | X | | I | Behavioral features, forum content and interaction features, temporal features | probabilistic soft logic (PSL); survival | Whether the student takes the last few | Reliable early prediction of student achievement using a latent model. |
| | | | | | | analysis | quizzes/assignments | |
| [93] | X | | | I | Content coverage, navigation (backjumps, textbook events) | multiple linear regression | Certificate (yes/no) | On average, certificate earners skip 22% of course content.<br><br>Certificate earners use non-linear navigation behaviour, often jumping backward to revisit earlier lectures. |

**Table 3: State-of-the-Art Research for Modelling, Predicting and influencing Learners' Achievement Based on Activities on the MOOC Platform and Background Information**

| Ref | Model? | Predict? | Influence? | Individual or collective? | Features used | Method used | Achievement metric | Findings |
|---|---|---|---|---|---|---|---|---|
| [11] | X | X | | I | lecture video views, quiz attempts, and discussion forum accesses<br><br>97 demographic features (from demographic survey, paying or not, geolocation information) | temporal interaction modeling techniques | Certificate of completion (yes/no) | Demographic information offers minimal predictive power compared to activity models, even when compared to models created very early on in the course<br><br>Adding demographics to activity models makes little difference. |
| [94] | X | X | | I | online resource use (time spent on homework, labs, lecture problems, lecture videos, tutorials, book, wiki, discussion board, score on first HW), student background (female, parent engineer, worked with other offline, teach EE, took diff. equations,) | Multiple regression methods | grade | Time spent on homework and time spent on the discussion board are related to higher scores.<br><br>Time spent on the book and the course wiki are related to lower scores.<br><br>Demographic factors such as gender are not related to scores.<br><br>Some background factors are strongly related to performance. Having studied the covered material previously and offline collaboration are related to higher scores. |

**Table 4: State-of-the-Art Research for Modelling, Predicting and influencing Learners' Achievement Based on Course and Platform Characteristics**

| Ref | Model? | Predict? | Influence? | Individual or collective? | Features used | Method used | Achievement metric | Findings |
|---|---|---|---|---|---|---|---|---|
| [95] | X | | | C | Due dates structure | statistics | Certificate (yes/no) | Stricter due dates are linked to higher certificate attainment rates.<br><br>Students who join late earn fewer certificates. |
| [96] | X | X | | I | Student course evaluation i.e. sentiment (for assignments, professor, discussion forum, course material), course characteristics (self-paced, difficulty, workload, weeks, certificate, peer assessments, team projects, suggested textbook, suggested paid textbook, final exam/project, interested users, course reviews) university characteristics (university ranking, courses offered), academic disciplines, platform characteristics, student characteristics (gender, attending university) | Grounded Theory Method (GTM) in a quantitative study<br><br>econometric, text mining, opinion mining, and machine learning techniques<br><br>latent regression model | Certificate (yes/no) | The more satisfied a student is with the professor, the teaching material, and the assignments, the more probable that s/he will successfully complete the course.<br><br>Learners satisfied with the discussion forum tend to complete the course less often.<br><br>All features except student characteristics help predict achievement. |
| [97] | | | X | I | Block ranking (completion of course activities) | gamification | N/A | (only proposed, no results yet) |
| [98] | | | X | I | no game elements (plain condition), game elements (game condition), social game elements (social condition) | Gamification<br><br>statistics | Videos watched, scores | A gamified course interface leads to an increase of 25% in retention period (videos watched) and 23% higher average scores.<br><br>Combining game elements with social elements is the most effective in retaining learners. |

**Table 5: Overview of MOOC Research based on the Section 2 AI and DM Taxonomy**

| | | AI | | | | | | | | | | | DM | | | |
|---|---|---|---|---|---|---|---|---|---|---|---|---|---|---|---|---|
| | | Problem Solving | | | Knowledge, reasoning and planning | | | Uncertain knowledge and reasoning | Learning | Communicating, perceiving and acting | | | Predictive | | Descriptive | |
| | | Searching | Game Theory | Constraint Satisfaction Problems | Knowledge engineering (propositional logic, first-order logic, inference) | Planning algorithms and solutions | Knowledge representation (ontologies, taxonomies, semantics, reasoning) | Knowledge and reasoning under uncertainty (traditional and Bayesian probability statistics, probabilistic reasoning, decision making) | Machine Learning | Communication algorithms and solutions (natural language processing | Perception algorithms and solutions (image and object recognition) | Robotics (perception, planning and moving) | interpolation and sequential prediction | Supervised Learning | Exploratory analysis | Clustering |
| Better Understanding Students | Engagement | | | | | | X | X | X | | | | | | | X |
| | Achievement | | X | | | | | X | X | X | | | | X | X | |
| | Knowledge | | | | | | | X | X | | | | | X | X | |
| Course contents | Videos | | | | | | X | | X | X | X | | | | X | X |
| | Supporting Materials | | | | | | X | | | | | | | | | |
| | Assessment | | | X | | X | X | X | X | | | | | X | | X |
| | Adaptive M&S | | | | X | | X | | X | | | | | X | | |
| | Authoring | | | | | | X | | X | | | | | | | |
| Community Building | Asynchronous Tools | | X | X | | | | | X | X | | | | X | X | X |
| | Synchronous Tools | | | | | | | | | | X | | | | X | |
| | Group Work | | | X | | | | | X | | | | | X | | |
| | Peer Support | | | | | | | | | | | | | | X | |
| Platform | Recommendation & Search | X | | | | | X | X | X | | | | | | X | |
| | Authentication | | | | | | | | X | X | X | | | | | |